\documentclass[conference]{IEEEtran}
\IEEEoverridecommandlockouts

\usepackage{cite}
\usepackage{amsmath,amssymb}
\usepackage{graphicx}
\usepackage{booktabs}
\usepackage{xcolor}
\usepackage{enumitem}
\usepackage[hidelinks]{hyperref}

\setlist[itemize]{leftmargin=1.1em,itemsep=1pt,topsep=2pt}

\begin{document}

\title{LADDER: Landmark-Only Lip-Sync Deepfake Detection Across Generators and Languages}

\author{\IEEEauthorblockN{Hao Chen}
\IEEEauthorblockA{haoc.research@gmail.com}
\and 
\IEEEauthorblockN{Junnan Xu}
\IEEEauthorblockA{junnanxu25@gmail.com}}

\maketitle

\begin{abstract}
We propose LADDER, a lip-sync deepfake detector that uses only the
trajectories of 64 lip landmarks. It needs no pixels and no audio. We find
two problems that make such a detector fail across domains. First, frame
rate is a confounding factor. On PolyGlotFake, real and fake videos differ
in frame rate, and the direction of this difference changes across
languages. We remove it by resampling every trajectory to 25\,fps. Second,
motion amplitude does not transfer across generators: some generators move
the mouth more than real speakers, and others move it less. The slope
across derivative orders does transfer, because it is higher in fake
videos for all five generators we test. LADDER combines a small trained
detector (0.31\,M parameters) with a prior score that needs no training:
the mean of the slope features, standardized on real videos only. Learning
this direction from data, or selecting it on the training domain, transfers
worse. LADDER runs at $7.2\times$ real time on one CPU core. It reaches
$0.917$ AUC on LipSyncTIMIT at CRF=23 and $0.850$ on seven languages of
PolyGlotFake. On clean video at CRF=23, the pixel-based LIPINC-V2 is
better ($0.943$). Under blur, noise, downscaling, low light and stronger
compression, LADDER is better. LADDER is also ahead of a retrained Xception in all 16 compression and corruption
settings, for example $0.887$ against $0.606$ under strong blur. Finally, a
decision threshold does not transfer across domains, but 20 real videos
from the target domain bring the false positive rate back to at most
$3.5\%$.
\end{abstract}

\section{Introduction}
\label{sec:intro}

Lip-sync deepfakes are hard to detect across generators. A lip-sync
generator changes only the mouth so that it matches new audio, so the fake
is cheap to make and hard to see. New generators appear quickly, and a
detector trained on one generator often fails on others. Three of the five
generators in LipSyncTIMIT~\cite{datta2025detecting}, for example, do not
appear in our training data.

Existing detectors need pixels or audio at test time. Most read the mouth
region as pixels, and others check whether the mouth matches the audio.
Both need sensitive data: face images or the soundtrack must reach the
detector. We ask how far we can go without either. Our detector, LADDER,
uses only the vertical trajectories of 64 lip landmarks. It is not the best
method on clean video. Under our protocol, LIPINC-V2~\cite{datta2025detecting}
is better at CRF=23, and AVAD~\cite{feng2023self} scores higher with audio in the tests of~\cite{datta2025detecting}. Two problems matter more than the network design.

\paragraph{Frame rate is a confounding factor.}
Features computed from frame differences depend on the frame rate, and a
window with a fixed number of frames covers a different length of time at
different frame rates. Most training data is 25\,fps, so the problem does
not show up in-domain. On PolyGlotFake~\cite{hou2024polyglotfake}, however,
real and fake videos have different shares of 30\,fps videos, and the
direction of the difference changes with the language. In French, a higher
frame rate means a real video. In Russian, it means a fake video. Without
correction, the per-language AUC follows this difference ($r=0.69$). We
resample every trajectory to 25\,fps before computing any feature, and the
correlation drops to $r=0.215$ (Sec.~\ref{sec:obs:fps}).

\begin{figure}[t]
\centering
\includegraphics[width=0.85\linewidth]{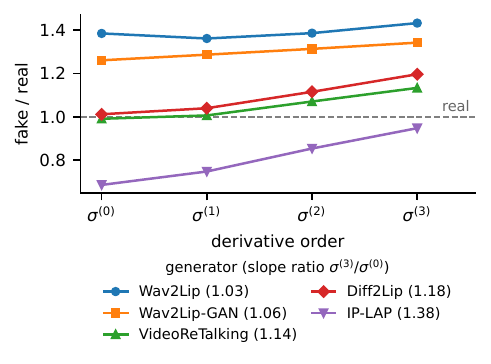}
\caption{Motion amplitude does not transfer across generators, but the
slope does. We plot the ratio of fake to real $\sigma^{(k)}$ for each
LipSyncTIMIT generator at CRF=23, after resampling to 25\,fps. At $k{=}0$
(amplitude), the generators lie on both sides of real speech (ratio 1).
Every curve is higher at $k{=}3$ than at $k{=}0$; the legend gives this
slope ratio $\sigma^{(3)}/\sigma^{(0)}$ of fakes over reals. IP-LAP has the
lowest amplitude and the steepest slope, and the trained detector fails on
it.}
\label{fig:teaser}
\end{figure}

\paragraph{Motion amplitude does not transfer across generators.}
It is often assumed that fake mouths move too much. This holds for Wav2Lip
and Wav2Lip-GAN. VideoReTalking and Diff2Lip move the mouth about as much
as real speakers, and IP-LAP~\cite{zhong2023identity} moves it less
(Figure~\ref{fig:teaser}). A detector that learns amplitude therefore scores
IP-LAP fakes as more real than real videos. The slope across derivative
orders does transfer. It describes how the motion variance changes from
position to velocity, acceleration and jerk, and it is higher in fake
videos for all five generators (Sec.~\ref{sec:obs:amp}).

LADDER combines a small trained detector with a prior score that needs no
training. The prior score is the mean of the slope features, standardized
on real videos only, and we fix it before we see any fake video. When the
same direction is learned from data or selected on the training domain, it
transfers worse (Sec.~\ref{sec:results:ablation}). LADDER is also robust to
image degradation. Under strong blur, a retrained Xception drops from
$0.883$ to $0.606$ AUC and LIPINC-V2 from $0.943$ to $0.821$, while LADDER
drops from $0.917$ to $0.887$ (Sec.~\ref{sec:results:robust}). Finally, AUC is
not enough for deployment. A threshold set at $1\%$ false positives on the
training domain gives $15$--$26\%$ false positives on LipSyncTIMIT and
$35$--$78\%$ across the PolyGlotFake languages. Twenty real videos from the
target domain fix this (Sec.~\ref{sec:deploy}).

\paragraph{Contributions.}
\begin{itemize}[leftmargin=*,itemsep=0pt,topsep=2pt]
  \item We show that frame rate is a confounding factor in landmark-based
        lip-sync detection whose direction changes across languages, and
        we remove it by resampling.
  \item We show that motion amplitude does not transfer across generators
        but the slope across derivative orders does, and we propose a
        training-free prior score based on it. Learning or selecting this
        direction from training data transfers worse.
  \item With no pixels or audio, LADDER is ahead of a retrained Xception in
        all 16 compression and corruption settings, and ahead of LIPINC-V2
        under blur, noise, downscaling and low light.
  \item We show that thresholds do not transfer across domains and propose
        a recalibration that needs only a few real videos from the target
        domain.
\end{itemize}

\section{Related work}
\label{sec:related}

\paragraph{Pixel-based detectors.}
Most lip-sync detectors read the mouth region as an image sequence.
LipForensics~\cite{haliassos2021lips} uses features from a lip-reading
network, RealForensics~\cite{haliassos2022leveraging} learns them from real
talking faces, and FTCN~\cite{zheng2021exploring} looks for temporal
inconsistency. LIPINC~\cite{datta2024exposing} compares mouth frames with
similar poses, both nearby and across the video. LIPINC-V2~\cite{datta2025detecting}
combines mouth crops with frame differences and gives the best reported
results on LipSyncTIMIT at CRF=23. It is also the strongest baseline in
our experiments (Sec.~\ref{sec:results:lst}).

\paragraph{Audio-visual detectors.}
Audio-visual detectors check whether the mouth matches the audio.
AVAD~\cite{feng2023self} fits a density to sync features of real videos,
Xiaolou Li et al.~\cite{li2024zero} compare the speech recognized from the audio and
from the lips,
LipFD~\cite{liu2024lips} compares video frames with the audio spectrogram, and
SpeechForensics~\cite{liang2024speechforensics} and
AVFF~\cite{oorloff2024avff} learn joint audio-visual features.
PIA~\cite{datta2025pia} matches phonemes to lip geometry and also tracks
changes of the face identity embedding. AVAD
gives the best reported result on LipSyncTIMIT at CRF=40~\cite{datta2025detecting}. All of them need
the audio at test time.

\paragraph{Landmark-based detectors.}
Landmark- and behavior-based detectors were mainly designed for face
swapping~\cite{agarwal2020detecting}. LRNet~\cite{sun2021improving} runs a
recurrent network on 68 facial landmarks and is trained on
FaceForensics++~\cite{rossler2019faceforensics++}. 
In these manipulations, the mouth still follows a real person. In lip-sync, the mouth motion itself is generated, so weights trained on these manipulations transfer poorly, as our results confirm.

\paragraph{Dataset shortcuts and one-class detection.} 
Smeu et al.~\cite{smeu2025circumventing} show that fake clips in FakeAVCeleb and
AV-Deepfake1M~\cite{cai2024av} start with a short silence that alone separates real and fake videos almost
perfectly, and propose to train on real data only. 
Yermakov et al.~\cite{yermakov2026deepfake} show that training on real and fake clips
from the same source video reduces shortcut learning. A split that
separates a fake from its source leaks in a similar way, so we split
AVLips by source video (Sec.~\ref{sec:setup}). Normalized lip
landmarks contain no background, but they do depend on the frame rate. To
our knowledge, frame rate has not been reported as a shortcut in this
setting. Training-free components also exist:
DIO~\cite{tan2024data} feeds fixed filters into a classifier. Our prior
score instead runs in parallel with the trained detector. Finally,
one-class detectors such as that of Bofan Chen et al.~\cite{Chen_2026_CVPR} learn only from
real videos and assume that real videos form a compact class. For lip
motion amplitude we find the opposite (Sec.~\ref{sec:obs:amp}).

\section{Preliminaries and two observations}
\label{sec:obs}

\subsection{Notation}
\label{sec:obs:notation}

For each frame $t$, we keep the normalized vertical coordinate $y_{i,t}$ of
lip landmark $i\in\{1,\dots,64\}$ (Sec.~\ref{sec:method:landmarks}). $\Delta$ is
the forward difference, $\Delta y_{i,t}=y_{i,t+1}-y_{i,t}$, and $\Delta^k$
approximates the $k$-th time derivative (velocity, acceleration and jerk for
$k=1,2,3$). In a window of $T$ frames, we compute
\begin{equation}
  \sigma_i^{(k)} = \operatorname{std}_{t}\big(\Delta^k y_{i,t}\big),
  \qquad k = 0,1,2,3 .
  \label{eq:sigma}
\end{equation}
We call $\big(\sigma_i^{(0)},\dots,\sigma_i^{(3)}\big)$ the
\emph{derivative ladder} of landmark $i$. $\sigma_i^{(0)}$ measures motion
amplitude, and higher orders measure how abruptly the motion changes.

\subsection{Observation 1: frame rate is a confounding factor}
\label{sec:obs:fps}

Frame rate $f$ changes the derivative ladder in two ways. First, a $k$-th
difference depends on the time step,
$\sigma^{(k)}\approx f^{-k}\operatorname{std}(y^{(k)})$. Second, a window of
$T{=}25$ frames lasts $1.00$\,s at 25\,fps but $0.83$\,s at 30\,fps, so it
covers a different part of each mouth movement. The second effect also
changes $\sigma^{(0)}$, so rescaling $\sigma^{(k)}$ by $(f/25)^k$ cannot
remove it. For $k{=}3$ and 30 to 25\,fps, the predicted factor is $1.728$,
but we measure $1.073$ on real LipSyncTIMIT videos, because the gain of a
$k$-th difference saturates near the Nyquist frequency. Simple rescaling
is therefore wrong.

Frame rate is linked to the label, and the direction of this link is not
stable. In LipSyncTIMIT, all real videos are 30\,fps, while
Diff2Lip~\cite{mukhopadhyay2024diff2lip} and IP-LAP output 25\,fps. Our
training data hides the problem, because $87.5\%$ of
AVLips~\cite{liu2024lips} and $98.8\%$ of
FakeAVCeleb~\cite{khalid2021fakeavceleb} are 25\,fps. On PolyGlotFake, the
real-minus-fake difference in the share of 30\,fps videos ranges from
$+36.3$ percentage points (French) to $-76.8$ (Russian). Because the sign
changes, no fixed correction per dataset or language can fix it. We
therefore resample every trajectory to one frame rate before computing any
feature (Sec.~\ref{sec:method:resample}).

\subsection{Observation 2: amplitude does not transfer, the slope does}
\label{sec:obs:amp}

Motion amplitude points in different directions for different generators.
Fig.~\ref{fig:teaser} shows the ratio of fake to real $\sigma^{(k)}$ for each
LipSyncTIMIT generator. At $k{=}0$, it ranges from $1.385$ (Wav2Lip) to
$0.686$ (IP-LAP). Wav2Lip and Wav2Lip-GAN move the mouth more than real
speakers, VideoReTalking~\cite{cheng2022videoretalking} and Diff2Lip match
them, and IP-LAP moves it less. A detector that learns amplitude therefore ranks IP-LAP fakes above
real videos. This does not depend on the architecture: we trained five
architectures with three feature sets, and all 15 models score IP-LAP below
chance at CRF=40.

The slope of the derivative ladder points in the same direction for all
generators. The fake-to-real ratio of $\sigma^{(3)}/\sigma^{(0)}$, which
removes any factor shared by all orders, lies between $1.034$ and $1.381$.
Wav2Lip~\cite{prajwal2020lip} raises all orders by a similar factor, so its
mouth simply moves too much. IP-LAP renders the mouth from landmarks that it predicts from the
audio~\cite{zhong2023identity}. Its amplitude is only $69\%$ of real
speech, but frame-to-frame jitter remains, so its slope is $38\%$ higher than in real
videos. All generators put too much energy into the higher orders relative
to the amplitude.

The slope does not come from the landmark extractor. We repeat the
measurement with a second landmark detector, FAN~\cite{bulat2017far}. The
amplitude ratio again lies on both sides of 1 ($0.78$ to $1.07$), and the
slope ratio is again above 1 for all five generators ($1.02$ to $1.43$).
The slope is also not a constant noise floor of the extractor. Such a floor
would raise the slope most for speakers who move little. Among real
LipSyncTIMIT videos, the slope instead rises with amplitude (Pearson
$r=0.34$).

The AVLips training split alone points to the same design. There, the
amplitude of real videos varies by a factor of $2.2$ between the 10th and
90th percentile, while the log ratios between orders vary only
$0.44$--$0.62$ times as much. Fake videos have $1.10$ times the real displacement but $1.17$--$1.19$
times the real higher orders, so their slope is already higher. In
domain, the slope is not the best single feature ($0.600$ AUC against
$0.668$ for $\log\sigma^{(1)}$). We choose it because it varies less across real
speakers.

Amplitude is also not suitable for one-class detection. Real speakers range
from calm to animated, while generators produce an average amount of
motion, so the real class is the wider one ($1.07$--$1.13$ times on the
AVLips training split). On LipSyncTIMIT at CRF=23, scoring the distance from the real-video mean
gives only $0.505$ AUC, and a Mahalanobis distance gives $0.485$. We fit
both on 98 real videos and test them on the other 104 real videos and all
fakes.

\section{Method}
\label{sec:method}

\begin{figure*}[t]
\centering
\includegraphics[width=\linewidth]{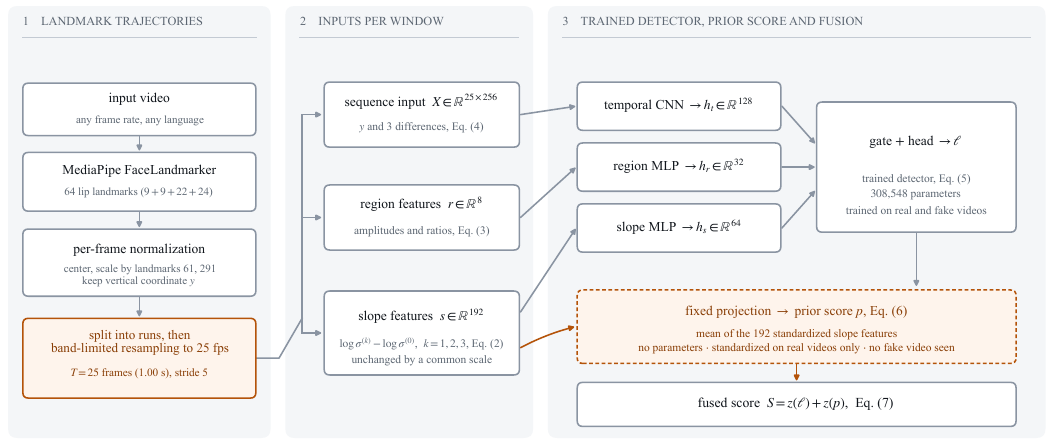}
\caption{Overview of LADDER. We extract 64 lip landmarks per frame,
resample the trajectories to 25\,fps and cut 25-frame windows
(Sec.~\ref{sec:method:resample}). Each window gives three inputs: the sequence
input $X$ (Eq.~\ref{eq:seq}), the region features $r$ (Eq.~\ref{eq:region}) and
the slope features $s$ (Eq.~\ref{eq:slope}). The trained detector reads all
three and outputs $\ell$ (Eq.~\ref{eq:detector}). The prior score $p$ reads
only $s$ and needs no training (Eq.~\ref{eq:prior}). The fused score $S$ adds
the two standardized scores (Eq.~\ref{eq:fusion}).}
\label{fig:pipeline}
\end{figure*}

LADDER extracts lip landmarks (Sec.~\ref{sec:method:landmarks}), resamples the
trajectories to 25\,fps (Sec.~\ref{sec:method:resample}) and computes three
inputs per window (Sec.~\ref{sec:method:features}). The trained detector reads
all three inputs (Sec.~\ref{sec:method:network}), the prior score reads only
the slope features (Sec.~\ref{sec:method:prior}), and the two scores are fused
(Sec.~\ref{sec:method:fusion}).

\subsection{Landmark trajectories}
\label{sec:method:landmarks}

We run MediaPipe FaceLandmarker~\cite{lugaresi2019mediapipe} on each frame
and keep 64 points around the mouth in four disjoint regions: lower outer
lip (9 points), lower inner lip (9), upper lip (22) and perioral region
(24). We will release the index lists with the code. We center each frame
on the midpoint of landmarks 61 and 291, scale it by their distance and
keep only the vertical coordinate. We do not fill
in frames where detection fails, because interpolation adds false
high-order variance. Instead, we split each video into runs of consecutive
frames and process each run separately.

\subsection{Resampling and windowing}
\label{sec:method:resample}

We resample each run to 25\,fps before windowing, with a band-limited
polyphase resampler (ratio $5/6$ for 30 and 29.97\,fps, $25/24$ for
24\,fps) and linear extension at the run ends. We choose 25\,fps because it
is the lowest frame rate common in our data. Upsampling to 30\,fps would
leave the band above 12.5\,Hz empty for 25\,fps sources, so energy in that
band would reveal the source frame rate instead of the label. We cut
windows of $T{=}25$ frames ($1.00$\,s) with a stride of 5.

\subsection{Inputs per window}
\label{sec:method:features}

\paragraph{Slope features $s$.}
Following Observation~2, we use log ratios between orders,
\begin{equation}
  s_{i,k} = \log \sigma_i^{(k)} - \log \sigma_i^{(0)},
  \qquad k=1,2,3,
  \label{eq:slope}
\end{equation}
which gives $s\in\mathbb{R}^{192}$. Multiplying a trajectory by a constant,
for example because of mouth size or camera distance, does not change $s$.

\paragraph{Region features $r$.}
Let $a_{\text{out}}, a_{\text{in}}, a_{\text{up}}, a_{\text{pe}}$ be the
mean $\sigma^{(0)}$ over the four regions. The region features compare
their motion:
\begin{equation}
  r=\Big(\tfrac{a_{\text{out}}}{a_{\text{up}}},
         \tfrac{a_{\text{in}}}{a_{\text{up}}},
         \tfrac{a_{\text{out}}}{a_{\text{in}}},
         \tfrac{a_{\text{pe}}}{a_{\text{up}}},
         \tfrac{a_{\text{out}}}{a_{\text{pe}}},
         \tfrac{a_{\text{out}}-a_{\text{up}}}{a_{\text{out}}+a_{\text{up}}},
         a_{\text{out}}, a_{\text{up}}\Big).
  \label{eq:region}
\end{equation}
The last two values are the only inputs that depend on scale. The trained
detector uses them, but the prior score does not.

\paragraph{Sequence input $X$.}
For each landmark, we stack the trajectory and its first three
differences, padded at the front to 25 frames:
\begin{equation}
  X=\big[\,y;\ \Delta y;\ \Delta^2 y;\ \Delta^3 y\,\big]\in\mathbb{R}^{25\times256},
  \label{eq:seq}
\end{equation}
that is, 25 frames $\times$ 64 landmarks $\times$ 4 orders. We compute
Eq.~\ref{eq:sigma} before padding.

\subsection{Trained detector}
\label{sec:method:network}

The trained detector has three branches. A 1-D CNN
($256\!\to\!128\!\to\!128\!\to\!64$ channels, kernel 3, multi-scale
pooling) maps $X$ to $h_t\in\mathbb{R}^{128}$. Two MLPs map $r$ to
$h_r\in\mathbb{R}^{32}$ and $s$ to $h_s\in\mathbb{R}^{64}$. A gate computes
$g=\operatorname{softmax}\big(\mathrm{MLP}([h_t;h_r;h_s])\big)\in\mathbb{R}^3$,
and a head outputs the detector score
\begin{equation}
  \ell = \mathrm{Head}\big([\,g_1 h_t;\; g_2 h_r;\; g_3 h_s\,]\big).
  \label{eq:detector}
\end{equation}
The model has $308{,}548$ parameters. We train it with binary cross-entropy,
AdamW~\cite{loshchilov2017decoupled} (learning rate $10^{-3}$, weight decay
$10^{-4}$), a cosine schedule over 60 epochs, batch size 512, a
class-balanced sampler and early stopping on validation AUC. Each of three
augmentations is applied with probability $0.5$: dropping and
interpolating one frame, scaling the amplitude by $[0.85,1.15]$, and
Gaussian noise ($\sigma=0.02$).

\subsection{Prior score}
\label{sec:method:prior}

The prior score is the mean of the standardized slope features. Let $\mu_j$
and $\tau_j$ be the mean and standard deviation of the $j$-th slope feature
over the windows of real training videos. Then
\begin{equation}
  p = \frac{1}{192}\sum_{j=1}^{192}\frac{s_j-\mu_j}{\tau_j}.
  \label{eq:prior}
\end{equation}
This is a projection onto the all-ones direction, and we use it for three
reasons. First, the direction comes from Observation~2, not from data: it
only says that a higher slope is suspicious. This holds for every generator
in Fig.~\ref{fig:teaser} and already on the AVLips training split. Second, it uses only real videos, so a user
can rebuild it from ordinary footage without any fake video. Third, a
learned direction transfers worse. The trained detector also sees $s$ but
does not find this direction, and learning the weight of $p$ end to end lowers the
mean AUC and the AUC on IP-LAP (Sec.~\ref{sec:results:ablation}).

\subsection{Fused score}
\label{sec:method:fusion}

Let $z(\cdot)$ standardize a score with its mean and standard deviation
over the windows of real training videos. The fused score of a window is
\begin{equation}
  S = z(\ell) + w\, z(p),\qquad w=1,
  \label{eq:fusion}
\end{equation}
and the score of a video is the mean over its windows. We fix $w{=}1$ in
advance and do not tune it.

\section{Experimental setup}
\label{sec:setup}

\paragraph{Datasets.}
Each train--test pair tests one type of shift.
\emph{AVLips $\rightarrow$ LipSyncTIMIT tests unseen generators.}
We use 6,981 AVLips~\cite{liu2024lips} videos made with
Wav2Lip~\cite{prajwal2020lip}, TalkLip~\cite{wang2023seeing},
SadTalker~\cite{zhang2023sadtalker} and MakeItTalk~\cite{zhou2020makelttalk}. 
We use the original-audio part of LipSyncTIMIT~\cite{datta2025detecting}:
202 real videos and 202 fake videos for each of five generators, at H.264
CRF=23 and 40. VideoReTalking, Diff2Lip and IP-LAP are unseen in training.
\emph{FakeAVCeleb $\rightarrow$ PolyGlotFake tests language and frame-rate
shift.} We use 718 real and 3,593 fake videos of
PolyGlotFake~\cite{hou2024polyglotfake} in seven languages. Its fakes
are made with Wav2Lip and VideoReTalking. Wav2Lip is also the lip-sync
generator of FakeAVCeleb~\cite{khalid2021fakeavceleb}, so this pair keeps
the generator roughly fixed. From FakeAVCeleb, we use the 500 real videos and the
15,872 fake videos with a Wav2Lip step. We drop the 4,694 videos made only
with FSGAN~\cite{nirkin2019fsgan} or FaceSwap~\cite{korshunova2017fast} and
the 500 with fake audio only, because their mouth motion comes from a real
speaker. A negative control confirms that these are label noise for our
method: the prior score separates them from real videos at only $0.552$
AUC, against $0.646$--$0.919$ for the five LipSyncTIMIT generators. We drop
videos outside $[20,50]$\,fps (none in LipSyncTIMIT, 3 in AVLips, 72 in
PolyGlotFake).

\paragraph{Protocol.}
We split AVLips by source video. Most AVLips fakes are made from a real
video of the same dataset. With a random split by video, $67\%$ of the
test videos have their source or a sibling fake in the training set
($73\%$ with clips of the same scene). AVLips has no source metadata, so we
match fakes to sources by file name (3,395 fakes) or thumbnail (181 fakes)
and group real clips of the same scene. The groups give 4,975 training,
976 validation and 1,030 test videos. All AVLips-trained models, including
the baselines, use this split. A random split changes the mean AUCs by at
most $0.013$ (supplementary material). No model sees LipSyncTIMIT or
PolyGlotFake during training, model selection or calibration. Real target
videos are used only in the one-class analysis (Sec.~\ref{sec:obs:amp}) and
the recalibration (Sec.~\ref{sec:deploy}).

\paragraph{Metric.}
We report video-level AUC, averaged over three seeds (42, 1, 22). The data
split does not depend on the seed, so all variants see the same test
videos. The supplementary material also reports average precision and
equal error rate.

\paragraph{Baselines.}
(i) \emph{Same protocol.} We retrain Xception~\cite{chollet2017xception}
and MesoInception4~\cite{afchar2018mesonet} on AVLips with our split and
the same three seeds. Both see the full face. We sample 10 frames per
video, detect the face with the OpenCV ResNet-10 SSD detector, enlarge the
box by $10\%$ and resize it ($299\times299$ for Xception, which starts from
ImageNet weights). We select the epoch by validation AUC, as for LADDER.
The video score is the mean frame probability.
(ii) \emph{Released weights.} LIPINC-V2~\cite{datta2025detecting} is
trained on FakeAVCeleb and reads RGB mouth crops. LipFD~\cite{liu2024lips} is trained on lip-sync fakes.
LRNet~\cite{sun2021improving} and
LipForensics~\cite{haliassos2021lips} are trained on FaceForensics++. We run all
of them with the authors' code on our test videos.
(iii) \emph{Reported results.} For AVAD~\cite{feng2023self}, we quote the
numbers that Datta et al.~\cite{datta2025detecting} report on LipSyncTIMIT. Their
protocol differs from ours, so they are not directly comparable. Our pipeline includes resampling, which
lowers our AUC (Sec.~\ref{sec:results:cost}); the baselines run on the
original frame rate.

\section{Results}
\label{sec:results}

In-domain AUC does not predict cross-domain performance. On the AVLips test
split, the trained detector reaches $0.988$ AUC, but only $0.698$ on
LipSyncTIMIT at CRF=40 and $0.334$ on IP-LAP.

\subsection{Across generators}
\label{sec:results:lst}

\begin{table*}[t]
\centering
\footnotesize
\setlength{\tabcolsep}{3pt}
\caption{Video-level AUC on LipSyncTIMIT. W2L: Wav2Lip, W2L-G:
Wav2Lip-GAN, VRT: VideoReTalking, D2L: Diff2Lip. $\ddagger$ marks
generators that do not appear in any training data. Unseen is the
mean over these three generators. AVLips-trained rows use our split and are means over three
seeds (s.d.\ over seeds: ours $\leq 0.029$, Xception $\leq 0.026$,
MesoInc.4 $\leq 0.065$). $\dagger$: trained on FakeAVCeleb. \textbf{Bold}: best per
column, excluding quoted results.}
\label{tab:main}
\begin{tabular}{@{}l cccccc c cccccc c@{}}
\toprule
& \multicolumn{7}{c}{CRF=23} & \multicolumn{7}{c}{CRF=40} \\
\cmidrule(lr){2-8}\cmidrule(l){9-15}
Method & W2L & W2L-G & VRT$^\ddagger$ & D2L$^\ddagger$ & IP-LAP$^\ddagger$ & Mean & Unseen
       & W2L & W2L-G & VRT$^\ddagger$ & D2L$^\ddagger$ & IP-LAP$^\ddagger$ & Mean & Unseen \\
\midrule
\multicolumn{15}{@{}l}{Same protocol: trained on AVLips with our split} \\
Xception~\cite{chollet2017xception}
  & .9747 & .9685 & .8039 & .8767 & .7910 & .8830 & .8239
  & .6919 & .6976 & .7210 & .6259 & .6029 & .6679 & .6500 \\
MesoInc.4~\cite{afchar2018mesonet}
  & .8369 & .8499 & .7788 & .7038 & .6804 & .7700 & .7210
  & .5811 & .6038 & .7333 & .5919 & .5962 & .6212 & .6405 \\
Ours: trained $\ell$
  & \textbf{.9822} & \textbf{.9796} & .9498 & .8760 & .6381 & .8851 & .8213
  & \textbf{.8898} & \textbf{.9130} & .7696 & .5829 & .3344 & .6980 & .5623 \\
Ours: prior $p$
  & .6459 & .6734 & .7468 & .7641 & \textbf{.9194} & .7499 & .8101
  & .5675 & .6304 & .5264 & \textbf{.6712} & \textbf{.8885} & .6568 & .6954 \\
Ours: fused $S$ (LADDER)
  & .9758 & .9766 & .9506 & .8915 & .7925 & .9174 & .8782
  & .8704 & .9004 & .7479 & .6234 & .5000 & \textbf{.7284} & .6238 \\
\midrule
\multicolumn{15}{@{}l}{Released weights, run on our test videos} \\
LIPINC-V2$^\dagger$~\cite{datta2025detecting}
  & .9811 & .9745 & \textbf{.9634} & \textbf{.8975} & .8960 & \textbf{.9425} & \textbf{.9190}
  & .7596 & .7376 & \textbf{.8401} & .6126 & .6817 & .7263 & \textbf{.7114} \\
LipFD~\cite{liu2024lips}
  & .6839 & .6956 & .7787 & .5257 & .5633 & .6494 & .6226
  & .5968 & .6181 & .7196 & .5017 & .5141 & .5901 & .5785 \\
LRNet~\cite{sun2021improving}
  & .5694 & .5323 & .6970 & .7001 & .7348 & .6467 & .7106
  & .4707 & .4351 & .4993 & .6104 & .5965 & .5224 & .5687 \\
LipForensics~\cite{haliassos2021lips}
  & .6600 & .6440 & .6951 & .5798 & .5563 & .6270 & .6104
  & .5946 & .5832 & .5594 & .5172 & .5182 & .5545 & .5316 \\
\midrule
\multicolumn{15}{@{}l}{Reported in~\cite{datta2025detecting} (different protocol)} \\
AVAD~\cite{feng2023self} (audio)
  & -- & -- & -- & -- & -- & .94 & --
  & -- & -- & -- & -- & -- & .81 & -- \\
\bottomrule
\end{tabular}
\end{table*}

\paragraph{The trained detector and the prior score fail on different
generators.}
The trained detector is close to perfect on Wav2Lip and Wav2Lip-GAN, which
are similar to its training data, but fails on IP-LAP (Table~\ref{tab:main}). At
CRF=40 its AUC there is $0.334$, below chance, as Observation~2 predicts.
The prior score is weakest on the two Wav2Lip variants at CRF=23
($0.646$ and $0.673$) and strongest on IP-LAP at both CRFs ($0.919$ and
$0.889$).

\paragraph{Fusion mainly fixes the worst case.}
The fused score improves the mean AUC by $0.032$ at CRF=23 and $0.030$ at
CRF=40. Most of the gain comes from IP-LAP ($+0.154$ and $+0.166$). At
CRF=40, however, fusion only brings IP-LAP back to chance ($0.500$). On the
other generators fusion costs at most $0.022$. We accept this cost, because
the generator is unknown at test time.

\paragraph{LADDER is ahead of Xception at both compression levels.}
LADDER reaches $0.917$ at CRF=23 against $0.883$ for Xception, and $0.728$
at CRF=40 against $0.668$. On the three unseen generators, LADDER is ahead
at CRF=23 ($0.878$ vs.\ $0.824$) but behind at CRF=40 ($0.624$ vs.\
$0.650$). The gap at CRF=40 comes from IP-LAP, where LADDER is at chance
and Xception reaches $0.603$. MesoInception4 and the other released-weight
baselines are far behind, including LRNet, which uses landmarks but is
trained on FaceForensics++.

\paragraph{LIPINC-V2 is the strongest baseline.}
LIPINC-V2 reaches $0.943$ at CRF=23 and $0.726$ at CRF=40, against $0.917$
and $0.728$ for LADDER. At CRF=23, almost all of the gap comes from IP-LAP
($0.896$ vs.\ $0.793$); on the other generators the two differ by at most
$0.013$. At CRF=40, LADDER is better on the two Wav2Lip variants, and
LIPINC-V2 on VideoReTalking and IP-LAP. LIPINC-V2 is ahead on the unseen
generators at both levels. AVAD scores higher in~\cite{datta2025detecting} but needs the audio.

\subsection{Across languages and training sets}
\label{sec:results:pgf}

\begin{table}[t]
\centering
\small
\setlength{\tabcolsep}{5pt}
\caption{AUC on PolyGlotFake (trained on FakeAVCeleb), sorted by
$\Delta$fps, the real-minus-fake difference in the share of 30\,fps videos
(percentage points). \textbf{Bold}: best per row.}
\label{tab:pgf}
\begin{tabular}{@{}lrccc@{}}
\toprule
Language & $\Delta$fps & Trained $\ell$ & Prior $p$ & Fused $S$ \\
\midrule
French   & $+36.3$ & .8308 & .7620 & \textbf{.8386} \\
Spanish  & $+35.7$ & .7977 & .7225 & \textbf{.8030} \\
Chinese  & $+32.6$ & .8577 & .8121 & \textbf{.8640} \\
Japanese & $+6.3$  & .9548 & .8429 & \textbf{.9595} \\
English  & $-23.0$ & .8601 & .8115 & \textbf{.8714} \\
Arabic   & $-51.7$ & .8129 & .6717 & \textbf{.8145} \\
Russian  & $-76.8$ & \textbf{.8069} & .5179 & .7984 \\
\midrule
Mean     &         & .8458 & .7344 & \textbf{.8499} \\
\bottomrule
\end{tabular}
\end{table}

\paragraph{LADDER reaches $0.850$ mean AUC over seven languages.}
With training on FakeAVCeleb, the fused AUC ranges from $0.798$ (Russian)
to $0.960$ (Japanese) (Table~\ref{tab:pgf}). Fusion adds only $0.004$ here,
because PolyGlotFake is closer to the training data: it also uses Wav2Lip. The dataset authors report $0.684$ with an audio-visual
model under a different protocol~\cite{hou2024polyglotfake}.

\paragraph{Resampling removes the frame-rate shortcut.}
Without resampling, the per-language AUC is correlated with $\Delta$fps
($r=0.69$). With resampling, the correlation of the trained detector drops
to $r=0.215$. The prior score is an exception ($r=0.711$) and fails on
Russian ($0.518$). There, resampling acts almost only on the fake videos and
removes the high-frequency motion that the prior score reads. The trained
detector is not affected, and the fused score stays close to it.

\paragraph{The other two pairings.}
Training on FakeAVCeleb instead of AVLips gives $0.925$ and $0.725$ on
LipSyncTIMIT, and the prior score changes by at most $0.002$ per generator,
because it only depends on statistics of real videos. On a subset of PolyGlotFake (supplementary material), the AVLips-trained
model reaches $0.927$, higher than the FakeAVCeleb-trained one, possibly
because AVLips contains four generators.

\subsection{Ablations}
\label{sec:results:ablation}

\begin{table}[t]
\centering
\small
\setlength{\tabcolsep}{4pt}
\caption{Ablations on LipSyncTIMIT (trained on AVLips, mean AUC over five
generators and three seeds). (a) Fusion weight $w$; we fix $w{=}1$ in
advance. (b) Direction of the prior score, sorted by AUC on the AVLips
training split; gain of the fused score over the trained detector. (c)
Fixed vs.\ learned use of the slope (IP: IP-LAP). \textbf{Bold}: best per
column.}
\label{tab:ablation}
\begin{tabular}{@{}lcc@{\hspace{14pt}}lcc@{}}
\multicolumn{6}{@{}l}{\textbf{(a) Fusion weight}} \\
\toprule
$w$ & @23 & @40 & $w$ & @23 & @40 \\
\midrule
0 (trained $\ell$) & .8851 & .6980 & 1.5 & .9229 & .7363 \\
0.25 & .8962 & .7073 & 2 & \textbf{.9238} & .7405 \\
0.5  & .9053 & .7156 & 3 & .9176 & \textbf{.7416} \\
0.75 & .9123 & .7224 & $\infty$ (prior $p$) & .7499 & .6568 \\
1 (ours) & .9174 & .7284 & & & \\
\bottomrule
\end{tabular}

\vspace{4pt}
\begin{tabular}{@{}lccc@{}}
\multicolumn{4}{@{}l}{\textbf{(b) Direction of the prior score}} \\
\toprule
Direction & In-domain & Gain @23 & Gain @40 \\
\midrule
$\Delta^1/\Delta^0$ only          & \textbf{.6651} & $+.0137$ & $+.0130$ \\
magnitude, all $\sigma^{(k)}$          & .6592 & $-.0335$ & $-.0182$ \\
all-ones (ours)                   & .6321 & $+.0323$ & $+.0305$ \\
$\Delta^2/\Delta^0$ only          & .6278 & $+.0328$ & $+.0316$ \\
$\Delta^3/\Delta^0$ only          & .5988 & $\mathbf{+.0464}$ & $\mathbf{+.0445}$ \\
\bottomrule
\end{tabular}

\vspace{4pt}
\begin{tabular}{@{}lccc@{}}
\multicolumn{4}{@{}l}{\textbf{(c) Fixed vs.\ learned use of the slope}} \\
\toprule
Variant & Mean @23 & IP @23 & IP @40 \\
\midrule
Trained detector $\ell$       & .8851 & .6381 & .3344 \\
\quad + learned weight on $p$ & .8702 & .5808 & .3119 \\
Network on $s$ only           & .8378 & .6843 & .3026 \\
Prior score $p$ (fixed)       & .7499 & \textbf{.9194} & \textbf{.8885} \\
Fused score $S$ (ours)        & \textbf{.9174} & .7925 & .5000 \\
\bottomrule
\end{tabular}
\end{table}

\paragraph{The fusion weight is not sensitive.}
Every $w$ between $0.25$ and $3$ improves over the trained detector
(Table~\ref{tab:ablation}a). The best test values ($w{=}2$ at CRF=23 and $w{=}3$
at CRF=40) would require tuning on the test set. Our fixed $w{=}1$ lies on
the same plateau.

\paragraph{Validation on the training domain picks the wrong direction.}
Ranked by AUC on the AVLips training split, the five candidate directions
for the prior score come out in almost the reverse order of their gains on
LipSyncTIMIT (Spearman $\rho=-0.9$, Table~\ref{tab:ablation}b). Magnitude is
the only direction that hurts across domains, although it ranks second in
domain. All slope-based directions help, so the key choice is to use
ratios between orders. The best ratio, $\Delta^3/\Delta^0$, would add
$0.014$ over ours, but it can only be found with test labels.

\paragraph{A learned direction transfers worse than the fixed one.}
Adding the prior score to the detector with a learnable weight lowers the
mean AUC from $0.885$ to $0.870$ and IP-LAP from $0.638$ to $0.581$
(Table~\ref{tab:ablation}c). A network trained on $s$ alone reaches only
$0.684$ and $0.303$ on IP-LAP, while the prior score on the same features
reaches $0.919$ and $0.889$. Inside the trained detector, $s$ also gives no
gain over raw $\sigma^{(0..3)}$ (difference $-0.002$ and $-0.011$). The
slope helps as a fixed prior, not as a learned feature.

\subsection{Robustness to compression and corruptions}
\label{sec:results:robust}

\begin{table}[t]
\centering
\small
\setlength{\tabcolsep}{3pt}
\caption{Robustness on LipSyncTIMIT (trained on AVLips, mean AUC over five
generators). Xception is
retrained under the same protocol (three seeds). LIPINC-V2 runs on the CPU,
so we test it on the original videos and five degraded settings only
(--: not tested). \textbf{Bold}: best per row.}
\label{tab:robust}
\begin{tabular}{@{}lccccc@{}}
\toprule
Condition & $\ell$ & $p$ & $S$ & Xception & LIPINC-V2 \\
\midrule
CRF 23, original    & .8851 & .7499 & .9174 & .8830 & \textbf{.9425} \\
CRF 23, re-encoded  & .8799 & .7508 & \textbf{.9154} & .8622 & -- \\
CRF 28              & .8620 & .7446 & \textbf{.9062} & .8190 & -- \\
CRF 32              & .8405 & .7371 & \textbf{.8888} & .7708 & .8018 \\
CRF 36              & .7907 & .7028 & \textbf{.8329} & .7313 & -- \\
CRF 40, re-encoded  & .6900 & .6765 & \textbf{.7291} & .6652 & -- \\
CRF 40, original    & .6980 & .6568 & \textbf{.7284} & .6679 & .7263 \\
\midrule
Blur $\sigma{=}1$   & .8775 & .7507 & \textbf{.9160} & .7957 & -- \\
Blur $\sigma{=}2$   & .8699 & .7233 & \textbf{.9072} & .6891 & -- \\
Blur $\sigma{=}3$   & .8509 & .6904 & \textbf{.8872} & .6062 & .8208 \\
Noise 10            & .8099 & .7020 & \textbf{.8775} & .7824 & -- \\
Noise 20            & .5397 & \textbf{.6842} & .6218 & .5496 & .6121 \\
Downscale $\times2$ & .8769 & .7479 & \textbf{.9126} & .8199 & -- \\
Downscale $\times4$ & .8633 & .7150 & \textbf{.8989} & .7448 & .8742 \\
Low light, mild     & .8623 & .7416 & \textbf{.9011} & .8436 & -- \\
Low light, strong   & .8529 & .7314 & \textbf{.8923} & .7603 & .5809 \\
\bottomrule
\end{tabular}
\end{table}

\paragraph{Setup.}
We test the models of Table~\ref{tab:main} on degraded copies of LipSyncTIMIT,
without retraining. We re-encode the CRF=23 videos at CRF 28 to 40, or
apply blur, pixel noise, downscaling or low light and re-encode at CRF=23
(details in the supplementary material). We extract the landmarks again on every degraded
copy. MediaPipe finds a face in at least $99.99\%$ of the frames, and the
landmarks move by at most $0.029$ mouth widths on average.

\paragraph{LADDER is ahead of Xception in every setting.}
LADDER has the higher AUC in all 16 settings (Table~\ref{tab:robust}). It is
also ahead of LIPINC-V2 in all five degraded settings where we test
LIPINC-V2. The gap to Xception is largest for blur, downscaling and low
light. Under strong blur, Xception drops from $0.883$ to $0.606$, while
LADDER drops from $0.917$ to $0.887$. A likely reason is that these
settings remove the fine texture a pixel detector relies on, while
MediaPipe still finds the lips. LIPINC-V2 is robust to blur ($0.821$) but
fails under strong low light ($0.581$, against $0.892$ for LADDER). Under
compression, LADDER degrades slowly up to CRF 32 ($0.889$) and faster
after it, while Xception drops to $0.771$ at CRF 32.

\paragraph{Noise is the hardest case, and the prior score is the most stable
part.}
At noise strength 20, the trained detector drops to $0.540$, Xception to
$0.550$ and LIPINC-V2 to $0.612$. Noise makes the landmarks jitter: it
raises $\sigma^{(3)}$ by a factor of $2.4$ in both real and fake videos.
The prior score stays at $0.684$, and between $0.89$ and $0.96$ on IP-LAP
in every setting. The gain of fusion grows from $0.036$ after re-encoding
to $0.082$ at noise strength 20.

\subsection{Cost of resampling}
\label{sec:results:cost}

Resampling lowers the mean AUC by $0.047$--$0.055$ at CRF=23 and by
$0.067$--$0.073$ at CRF=40 (paired over seeds, two architectures, random
split). Most of the loss falls on the three generators that output 30\,fps
like the real videos. There, resampling removes useful motion between
$12.5$ and $15$\,Hz. This is the price of accepting any input frame rate.

\section{Deployment}
\label{sec:deploy}

A deployed detector needs a fixed threshold, so we also report false
positive rates (FPR). System A is trained on AVLips, calibrated on its real
validation videos and tested on LipSyncTIMIT. System B is trained on
FakeAVCeleb and tested on PolyGlotFake; its threshold is set on windows, so
its FPR is a lower bound.

\paragraph{A threshold does not transfer across domains.}
A threshold set for $1\%$ FPR on the calibration data gives $15.2\%$ and
$25.7\%$ FPR on LipSyncTIMIT at CRF=23 and 40, and $34.8$--$77.6\%$ across
the PolyGlotFake languages (Table~\ref{tab:recal}). The ranking does not change,
but the scores of real videos shift. The mean fused score of real videos
is 0 on the training domain by construction, but $2.1$ on LipSyncTIMIT at
CRF=23 and $2.8$ at CRF=40. On System A, fusion lowers the FPR from
$30.5\%$ to $15.2\%$, because the prior score depends only on real videos.

\begin{table}[t]
\centering
\small
\setlength{\tabcolsep}{4pt}
\caption{FPR of the fused score at a threshold set for $1\%$, used
unchanged or recalibrated on $N$ real target videos. A: AVLips
$\rightarrow$ LipSyncTIMIT (mean over 20 random draws of the $N$ videos;
range over the five generator subsets). B: FakeAVCeleb $\rightarrow$
PolyGlotFake (range over languages; $N{=}50$ covers the six languages with
enough real videos).}
\label{tab:recal}
\begin{tabular}{@{}lcccc@{}}
\toprule
Target & Unchanged & $N{=}10$ & $N{=}20$ & $N{=}50$ \\
\midrule
A, CRF=23 & .152 & .017--.032 & .010--.013 & .006--.008 \\
A, CRF=40 & .257 & .016--.023 & .007--.010 & .005--.008 \\
B         & .348--.776 & .005--.072 & .000--.035 & .000--.028 \\
\bottomrule
\end{tabular}
\end{table}

\paragraph{A few real videos fix the threshold.}
We re-estimate the mean and standard deviation of the score from $N$ real
videos of the target domain and keep the threshold at the same number of
standard deviations. No labels or fake videos are needed. With $N{=}20$,
the FPR drops to at most $3.5\%$, and with $N{=}50$ to at most $2.8\%$. On
System A, $N{=}20$ already gives $0.7$--$1.3\%$. The true positive rate
must be read with it. With $N{=}50$, LADDER detects $13\%$ of the
Diff2Lip fakes and $2\%$ of the IP-LAP fakes at CRF=23, although its AUC
on IP-LAP is $0.793$.

\paragraph{LADDER is cheap and sends little data.}
On one core of an AMD Ryzen 7 9700X, LADDER processes one second of video
in $138$\,ms ($7.2\times$ real time). Landmark extraction takes $98\%$ of
this time. The detector needs only 64 numbers per frame. This reduces the
data that leaves the device, but it is not anonymous, because mouth motion
can identify a person.

\section{Limitations and conclusion}
\label{sec:limitations}

\paragraph{Limitations.}
Heavy compression and strong noise are the hardest cases. Compression
removes high-frequency motion. At CRF=40, the fake-to-real slope ratio
shrinks to $0.98$--$1.31$, and it falls below 1 for Wav2Lip and
VideoReTalking. The fused AUC is then $0.728$, and only $0.500$ on IP-LAP.
Strong noise adds landmark jitter to both real and fake videos, and the
fused AUC on IP-LAP drops to $0.366$. LADDER degrades less than Xception,
but we have no fix. On clean video, the pixel-based LIPINC-V2 is better,
mainly on IP-LAP. The slope signal holds for two landmark extractors, but
we train the detector only with MediaPipe, and we did not test occlusion
or large head pose. Upsampled videos keep a weaker frame-rate effect
($r\approx0.35$). Our split recovers the AVLips sources from file names and
thumbnails. We found no source for 630 fakes, so some leakage through them
may remain. The FakeAVCeleb training set has one generator family, our
error bars cover three seeds, and the baselines are trained without image
corruptions, so our robustness comparison holds for the same protocol
only. Recalibration fixes the false positive rate but not the recall
(Sec.~\ref{sec:deploy}).
Finally, a generator with a smoothness loss on higher orders may weaken
the slope signal. IP-LAP already produces low-amplitude motion and is
still detected by the prior score, but we expect this margin to shrink.

\paragraph{Broader impact.}
Without recalibration, the FPR ranges from $34.8\%$ to $77.6\%$ across
languages, so speakers of some languages would be falsely flagged much more
often. We recommend recalibrating on local real videos before deployment.
All datasets we use are publicly available for research.

\paragraph{Conclusion.}
Lip-sync deepfakes can be detected from lip landmark trajectories alone.
Frame rate must be normalized on the trajectories, because its link to the
label changes across languages. The decision should use the slope across
derivative orders, because amplitude changes in different directions for
different generators. A fixed prior score along this slope complements the
trained detector and transfers better than a learned or selected
direction. LADDER is close to the best pixel-based detector on clean video and more
robust than Xception and LIPINC-V2 under image degradation.

\bibliographystyle{IEEEtran}
\bibliography{refs}

@inproceedings{haliassos2021lips,
  title={Lips don't lie: A generalisable and robust approach to face forgery detection},
  author={Haliassos, Alexandros and Vougioukas, Konstantinos and Petridis, Stavros and Pantic, Maja},
  booktitle={2021 IEEE/CVF Conference on Computer Vision and Pattern Recognition (CVPR)},
  pages={5039--5049},
  year={2021}
}

@inproceedings{zheng2021exploring,
  title={Exploring temporal coherence for more general video face forgery detection},
  author={Zheng, Yinglin and Bao, Jianmin and Chen, Dong and Zeng, Ming and Wen, Fang},
  booktitle={2021 IEEE/CVF International Conference on Computer Vision (ICCV)},
  pages={15024--15034},
  year={2021}
}

@article{datta2025detecting,
  title={Detecting lip-syncing deepfakes: Vision temporal transformer for analyzing mouth inconsistencies},
  author={Datta, Soumyya Kanti and Jia, Shan and Lyu, Siwei},
  journal={arXiv preprint arXiv:2504.01470},
  year={2025}
}

@inproceedings{chollet2017xception,
  title={Xception: Deep learning with depthwise separable convolutions},
  author={Chollet, Fran{\c{c}}ois},
  booktitle={2017 {IEEE} Conference on Computer Vision and Pattern Recognition, {CVPR}},
  pages={1800--1807},
  year={2017}
}

@inproceedings{afchar2018mesonet,
  title={{M}eso{N}et: a compact facial video forgery detection network},
  author={Afchar, Darius and Nozick, Vincent and Yamagishi, Junichi and Echizen, Isao},
  booktitle={2018 {IEEE} International Workshop on Information Forensics and Security, {WIFS}},
  pages={1--7},
  year={2018}
}

@inproceedings{liu2024lips,
  title={Lips are lying: Spotting the temporal inconsistency between audio and visual in lip-syncing deepfakes},
  author={Liu, Weifeng and She, Tianyi and Liu, Jiawei and Li, Boheng and Yao, Dongyu and Liang, Ziyou and Wang, Run},
  booktitle    = {Advances in Neural Information Processing Systems 37: Annual Conference
                  on Neural Information Processing Systems 2024, NeurIPS},
  year={2024}
}

@inproceedings{feng2023self,
  title={Self-supervised video forensics by audio-visual anomaly detection},
  author={Feng, Chao and Chen, Ziyang and Owens, Andrew},
  booktitle={2023 IEEE/CVF Conference on Computer Vision and Pattern Recognition (CVPR)},
  pages={10491--10503},
  year={2023}
}

@inproceedings{li2024zero,
  author       = {Xiaolou Li and
                  Zehua Liu and
                  Chen Chen and
                  Lantian Li and
                  Li Guo and
                  Dong Wang},
  title        = {Zero-Shot Fake Video Detection by Audio-Visual Consistency},
  booktitle    = {25th Annual Conference of the International Speech Communication Association,
                  Interspeech},
  year         = {2024}
}

@inproceedings{sun2021improving,
  author       = {Zekun Sun and
                  Yujie Han and
                  Zeyu Hua and
                  Na Ruan and
                  Weijia Jia},
  title        = {Improving the Efficiency and Robustness of Deepfakes Detection Through
                  Precise Geometric Features},
  booktitle    = {{IEEE} Conference on Computer Vision and Pattern Recognition, {CVPR}},
  pages        = {3609--3618},
  year         = {2021}
}

@inproceedings{smeu2025circumventing,
  title={Circumventing shortcuts in audio-visual deepfake detection datasets with unsupervised learning},
  author={Smeu, Stefan and Boldisor, Dragos-Alexandru and Oneata, Dan and Oneata, Elisabeta},
  booktitle={2025 IEEE/CVF Conference on Computer Vision and Pattern Recognition (CVPR)},
  pages={18815--18825},
  year={2025}
}

@article{tan2024data,
  title={Data-independent operator: A training-free artifact representation extractor for generalizable deepfake detection},
  author={Tan, Chuangchuang and Liu, Ping and Tao, RenShuai and Liu, Huan and Zhao, Yao and Wu, Baoyuan and Wei, Yunchao},
  journal={arXiv preprint arXiv:2403.06803},
  year={2024}
}

@inproceedings{Chen_2026_CVPR,
  author       = {Bofan Chen and
                  Hongyu Zhu and
                  Yi He and
                  Sichu Liang and
                  Shi{-}Lin Wang},
  title        = {Learning Forgery-Aware Lip Representations Without Forgery Priors},
  booktitle    = {{IEEE/CVF} Conference on Computer Vision and Pattern Recognition,
                  {CVPR}},
  pages        = {42911--42921},
  year         = {2026}
}

@inproceedings{agarwal2020detecting,
  title={Detecting deep-fake videos from appearance and behavior},
  author={Agarwal, Shruti and Farid, Hany and El-Gaaly, Tarek and Lim, Ser-Nam},
  booktitle={12th {IEEE} International Workshop on Information Forensics and Security, {WIFS}},
  pages={1--6},
  year={2020}
}

@inproceedings{prajwal2020lip,
  title={A lip sync expert is all you need for speech to lip generation in the wild},
  author={Prajwal, KR and Mukhopadhyay, Rudrabha and Namboodiri, Vinay P and Jawahar, CV},
  booktitle={{MM} '20: The 28th {ACM} International Conference on Multimedia},
  pages={484--492},
  year={2020}
}

@inproceedings{cheng2022videoretalking,
  title={{V}ideo{R}e{T}alking: Audio-based lip synchronization for talking head video editing in the wild},
  author={Cheng, Kun and Cun, Xiaodong and Zhang, Yong and Xia, Menghan and Yin, Fei and Zhu, Mingrui and Wang, Xuan and Wang, Jue and Wang, Nannan},
  booktitle={SIGGRAPH Asia 2022 Conference Papers},
  pages={30:1--30:9},
  year={2022}
}

@inproceedings{mukhopadhyay2024diff2lip,
  title={{D}iff2{L}ip: Audio conditioned diffusion models for lip-synchronization},
  author={Mukhopadhyay, Soumik and Suri, Saksham and Gadde, Ravi Teja and Shrivastava, Abhinav},
  booktitle={2024 IEEE/CVF Winter Conference on Applications of Computer Vision (WACV)},
  pages={5280--5290},
  year={2024}
}

@inproceedings{zhong2023identity,
  title={Identity-preserving talking face generation with landmark and appearance priors},
  author={Zhong, Weizhi and Fang, Chaowei and Cai, Yinqi and Wei, Pengxu and Zhao, Gangming and Lin, Liang and Li, Guanbin},
  booktitle={2023 IEEE/CVF Conference on Computer Vision and Pattern Recognition (CVPR)},
  pages={9729--9738},
  year={2023}
}

@inproceedings{nirkin2019fsgan,
  title={{FSGAN}: Subject agnostic face swapping and reenactment},
  author={Nirkin, Yuval and Keller, Yosi and Hassner, Tal},
  booktitle={2019 {IEEE/CVF} International Conference on Computer Vision, {ICCV}},
  pages={7183--7192},
  year={2019}
}

@inproceedings{korshunova2017fast,
  title={Fast face-swap using convolutional neural networks},
  author={Korshunova, Iryna and Shi, Wenzhe and Dambre, Joni and Theis, Lucas},
  booktitle={2017 IEEE International Conference on Computer Vision (ICCV)},
  pages={3697--3705},
  year={2017}
}

@inproceedings{khalid2021fakeavceleb,
  author       = {Hasam Khalid and
                  Shahroz Tariq and
                  Minha Kim and
                  Simon S. Woo},
  title        = {{FakeAVCeleb}: {A} Novel Audio-Video Multimodal Deepfake Dataset},
  booktitle    = {Proceedings of the Neural Information Processing Systems Track on
                  Datasets and Benchmarks 1, NeurIPS Datasets and Benchmarks},
  year         = {2021}
}

@inproceedings{hou2024polyglotfake,
  title={{P}oly{G}lot{F}ake: A novel multilingual and multimodal deepfake dataset},
  author={Hou, Yang and Fu, Haitao and Chen, Chunkai and Li, Zida and Zhang, Haoyu and Zhao, Jianjun},
  booktitle={Pattern Recognition - 27th International Conference, {ICPR}},
  volume       = {15314},
  pages={180--193},
  year={2024}
}

@inproceedings{rossler2019faceforensics++,
  title={{F}ace{F}orensics++: Learning to detect manipulated facial images},
  author={Andreas R{\"{o}}ssler and
                  Davide Cozzolino and
                  Luisa Verdoliva and
                  Christian Riess and
                  Justus Thies and
                  Matthias Nie{\ss}ner},
  booktitle={2019 {IEEE/CVF} International Conference on Computer Vision, {ICCV}},
  pages={1--11},
  year={2019}
}

@article{lugaresi2019mediapipe,
  title={{M}edia{P}ipe: A framework for building perception pipelines},
  author={Camillo Lugaresi and
                  Jiuqiang Tang and
                  Hadon Nash and
                  Chris McClanahan and
                  Esha Uboweja and
                  Michael Hays and
                  Fan Zhang and
                  Chuo{-}Ling Chang and
                  Ming Guang Yong and
                  Juhyun Lee and
                  Wan{-}Teh Chang and
                  Wei Hua and
                  Manfred Georg and
                  Matthias Grundmann},
  journal={arXiv preprint arXiv:1906.08172},
  year={2019}
}

@inproceedings{wang2023seeing,
  title={Seeing what you said: Talking face generation guided by a lip reading expert},
  author={Wang, Jiadong and Qian, Xinyuan and Zhang, Malu and Tan, Robby T and Li, Haizhou},
  booktitle={2023 IEEE/CVF Conference on Computer Vision and Pattern Recognition (CVPR)},
  pages={14653--14662},
  year={2023}
}

@inproceedings{haliassos2022leveraging,
  title={Leveraging real talking faces via self-supervision for robust forgery detection},
  author={Haliassos, Alexandros and Mira, Rodrigo and Petridis, Stavros and Pantic, Maja},
  booktitle={2022 IEEE/CVF Conference on Computer Vision and Pattern Recognition (CVPR)},
  pages={14930--14942},
  year={2022}
}

@inproceedings{oorloff2024avff,
  title={{AVFF}: Audio-visual feature fusion for video deepfake detection},
  author={Oorloff, Trevine and Koppisetti, Surya and Bonettini, Nicol{\`o} and Solanki, Divyaraj and Colman, Ben and Yacoob, Yaser and Shahriyari, Ali and Bharaj, Gaurav},
  booktitle={2024 IEEE/CVF Conference on Computer Vision and Pattern Recognition (CVPR)},
  pages={27092--27102},
  year={2024}
}

@inproceedings{datta2025pia,
  title={{PIA:} Deepfake Detection Using Phoneme-Temporal and Identity-Dynamic Analysis},
  author={Datta, Soumyya Kanti and Ranga, Tanvi and Sun, Chengzhe and Lyu, Siwei},
  booktitle={{IEEE/CVF} International Conference on Computer Vision, {ICCV} 2025 - Workshops},
  pages={1607--1617},
  year={2025}
}

@inproceedings{datta2024exposing,
  title={Exposing lip-syncing deepfakes from mouth inconsistencies},
  author={Datta, Soumyya Kanti and Jia, Shan and Lyu, Siwei},
  booktitle={2024 IEEE International Conference on Multimedia and Expo (ICME)},
  pages={1--6},
  year={2024}
}

@inproceedings{zhang2023sadtalker,
  title={{S}ad{T}alker: Learning realistic 3{D} motion coefficients for stylized audio-driven single image talking face animation},
  author={Zhang, Wenxuan and Cun, Xiaodong and Wang, Xuan and Zhang, Yong and Shen, Xi and Guo, Yu and Shan, Ying and Wang, Fei},
  booktitle={2023 IEEE/CVF Conference on Computer Vision and Pattern Recognition (CVPR)},
  pages={8652--8661},
  year={2023}
}

@article{zhou2020makelttalk,
  title={{M}ake{I}t{T}alk: {S}peaker-aware talking-head animation},
  author={Zhou, Yang and Han, Xintong and Shechtman, Eli and Echevarria, Jose and Kalogerakis, Evangelos and Li, Dingzeyu},
  journal={ACM Transactions On Graphics (TOG)},
  volume={39},
  number={6},
  pages={1--15},
  year={2020}
}

@inproceedings{cai2024av,
  title={{AV}-{D}eepfake1{M}: A large-scale {LLM}-driven audio-visual deepfake dataset},
  author={Cai, Zhixi and Ghosh, Shreya and Adatia, Aman Pankaj and Hayat, Munawar and Dhall, Abhinav and Gedeon, Tom and Stefanov, Kalin},
  booktitle={Proceedings of the 32nd {ACM} International Conference on Multimedia, {MM}},
  pages={7414--7423},
  year={2024}
}

@inproceedings{loshchilov2017decoupled,
  author       = {Ilya Loshchilov and
                  Frank Hutter},
  title        = {Decoupled Weight Decay Regularization},
  booktitle    = {7th International Conference on Learning Representations, {ICLR}},
  year         = {2019}
}

@inproceedings{yermakov2026deepfake,
  title={Deepfake detection that generalizes across benchmarks},
  author={Yermakov, Andrii and Cech, Jan and Matas, Jiri and Fritz, Mario},
  booktitle={2026 IEEE/CVF Winter Conference on Applications of Computer Vision (WACV)},
  pages={773--783},
  year={2026}
}

@inproceedings{liang2024speechforensics,
  title={{SpeechForensics}: Audio-visual speech representation learning for face forgery detection},
  author={Liang, Yachao and Yu, Min and Li, Gang and Jiang, Jianguo and Li, Boquan and Yu, Feng and Zhang, Ning and Meng, Xiang and Huang, Weiqing},
  booktitle    = {Advances in Neural Information Processing Systems 37: Annual Conference
                  on Neural Information Processing Systems 2024, NeurIPS},
  year={2024}
}

@inproceedings{bulat2017far,
  title={How Far are We from Solving the {2D} {\&} {3D} Face Alignment Problem? (and a Dataset of 230, 000 {3D} Facial Landmarks)},
  author={Bulat, Adrian and Tzimiropoulos, Georgios},
  booktitle={{IEEE} International Conference on Computer Vision, {ICCV}},
  pages={1021--1030},
  year={2017}
}

\end{document}